\documentclass[10pt, conference, letterpaper]{IEEEtran}
\usepackage{url}
\usepackage[utf8]{inputenc}
\usepackage{xcolor}
\usepackage{amsmath}
\usepackage{multirow}
\usepackage{amssymb}
\usepackage{enumitem}
\usepackage{bm}
\usepackage{graphicx}

\usepackage{gensymb}

\usepackage[acronyms,nonumberlist,nopostdot,nomain,nogroupskip]{glossaries}
\usepackage{tablefootnote}
\usepackage{booktabs}
\usepackage{tabularx}
\usepackage{epsfig}
\usepackage[outdir=images_final/]{epstopdf}
\usepackage{pgfplots}
\pgfplotsset{compat=newest} 
\pgfplotsset{plot coordinates/math parser=false} 
\newlength\fheight
\newlength\fwidth
\usetikzlibrary{plotmarks,patterns,decorations.pathreplacing,backgrounds,calc,arrows,arrows.meta,spy,matrix}
\usepgfplotslibrary{patchplots,groupplots}
\usepackage{tikzscale}
\usepackage{siunitx}

\usepackage{multirow}

\usepackage{multirow}

\renewcommand{\figurename}{Fig.}
\usepackage[font=scriptsize]{subcaption}
\usepackage[font=footnotesize]{caption}

\usepackage{mathtools}

\usepackage{dblfloatfix}    
\usepackage{colortbl}

\newacronym{iot}{IoT}{Internet of Things}
\newacronym{ml}{ML}{Machine Learning}
\newacronym{dl}{DL}{Deep Learning}
\newacronym{marl}{MARL}{Multi-Agent Reinforcement Learning}
\newacronym{rl}{RL}{Reinforcement Learning}
\newacronym{decpomdp}{Dec-POMDP}{Decentralized Partially Observable Markov Decision Process }
\newacronym{pomdp}{POMDP}{Partially Observable Markov Decision Process}
\newacronym{uav}{UAV}{Unmanned Aerial Vehicle}
\newacronym{dqn}{DQN}{Deep Q-Network}
\newacronym{dnn}{DNN}{Deep Neural Network}
\newacronym{dial}{DIAL}{Differentiable Inter-Agent Learning}
\newacronym{mdp}{MDP}{Markov Decision Process}
\newacronym{fov}{FoV}{Field of View}
\newacronym{cnn}{CNN}{Convolutional Neural Network}
\newacronym{nn}{NN}{Neural Network}
\newacronym{ddql}{DDQL}{Distributed Deep Q-Learning}
\newacronym{pdf}{PDF}{Probability Density Function}
\newacronym{ndpomdp}{ND-POMDP}{Networked Distributed Partially Observable Markov Decision Process}
\newacronym{radam}{RAdam}{Rectified Adam}
\newacronym{cdf}{CDF}{Cumulative Distribution Function}
\newacronym{mpc}{MPC}{Model Predictive Control}
\newacronym{rv}{rv}{Random Variable}
\newacronym{qoe}{QoE}{Quality of Experience}
\newacronym{tlc}{TLC}{Telecommunications}
\newacronym{cml}{CML}{Communications for Machine Learning}
\newacronym{mlc}{MLC}{Machine Learning for Communications}
\newacronym{drl}{DRL}{Deep Reinforcement Learning}
\newacronym{rf}{RF}{Radio Frequency}
\newacronym{urllc}{URLLC}{Ultra-Reliable and Low-Latency Communications}
\newacronym{fl}{FL}{federated learning}
\newacronym{kpi}{KPI}{Key Performance Indicators}
\newacronym{mec}{MEC}{Mobile Edge Computing}
\newacronym{ei}{EI}{Edge Intelligence}
\newacronym{bs}{BS}{base station}
\newacronym{sdn}{SDN}{Software Defined Networking}
\newacronym{mimo}{MIMO}{Multiple-Input Multiple-Output}
\newacronym{gp}{GP}{Gaussian Process}
\newacronym{iiot}{IIoT}{Industrial Internet of Things}
\newacronym{csi}{CSI}{Channel State Information}
\newacronym{sgd}{SGD}{Stochastic Gradient Descent}
\newacronym{iid}{iid}{independent and identically distributed}
\newacronym{ofdm}{OFDM}{Orthogonal Frequency Division Multiplexing}
\newacronym{los}{LOS}{Line-of-Sight}
\newacronym{nlos}{NLOS}{Non-Line-of-Sight}
\newacronym{snr}{SNR}{Signal to Noise Ratio}
\newacronym{rb}{RB}{Resource Block}
\newacronym{6g}{6G}{sixth generation}
\newacronym{ai}{AI}{artificial intelligence}
\newacronym{sfl}{SFL}{Synchronous Federated Learning}
\newacronym{frfl}{FRFL}{Fixed Rate Federated Learning}

\usepackage[capitalize]{cleveref}
\definecolor{steelblue}{RGB}{176,196,222}

\makeglossaries
\begin{document}

	
	\title{On the Convergence Time of Federated Learning Over Wireless Networks Under Imperfect CSI}
	
	\author{Francesco Pase, Marco Giordani, Michele Zorzi \\ Department of Information Engineering, University of Padova, Padova, Italy, and WiLab/CNIT \\ Email: \texttt{\{name.surname\}@dei.unipd.it} }

	\maketitle

	\begin{abstract}
		Federated learning (FL) has recently emerged as an attractive decentralized solution for wireless networks to collaboratively train a shared model while keeping data localized. 
		As a general approach, existing FL methods tend to assume perfect knowledge of the Channel State Information (CSI) during the training phase, which may not be easy to acquire in case of fast fading channels. Moreover, literature analyses either consider a fixed number of clients participating in the training of the federated model, or simply assume that all clients operate at the maximum achievable rate to transmit model data.
		In this paper, we fill these gaps by proposing a training process that takes channel statistics as a bias to minimize the convergence time under imperfect CSI.
		Numerical experiments demonstrate that it is possible to reduce the training time by neglecting model updates from clients that cannot sustain a minimum predefined transmission rate. 
		We also examine the trade-off between number of clients involved in the training process and model accuracy as a function of different fading regimes.
	\end{abstract}
		\begin{tikzpicture}[remember picture,overlay]
\node[anchor=north,yshift=-10pt] at (current page.north) {\parbox{\dimexpr\textwidth-\fboxsep-\fboxrule\relax}{
\centering\footnotesize This paper has been accepted for presentation at the IEEE ICC 2021 Workshop on Edge Learning for 5G Mobile Networks and Beyond. \textcopyright 2021 IEEE.\\
Please cite it as: F. Pase, M. Giordani, M. Zorzi, “On the Convergence Time of Federated Learning Over Wireless Networks Under Imperfect CSI,” \\ IEEE International Conference on Communications Workshops (ICC WKSHPS), Virtual/Montreal, Canada, 2021.}};
\end{tikzpicture}
	
	\section{Introduction}
	\label{sec:intro}
	\Gls{ai}  will play a more and more prominent role in the design and optimization of \gls{6g} wireless networks~\cite{giordani2020toward}.
	Notably, it is envisioned that the co-design of communications systems and applications running on top of them will facilitate an efficient use of wireless physical resources, thereby enabling future vertical services to fulfill very demanding sets of requirements.
	In particular, \emph{\gls{fl}} has gained a lot of interest as a promising and efficient tool to bring intelligence to the edge, where devices collaborate to maintain fresh learning models rather than uploading raw data to centralized servers~\cite{BrendanMcMahan2017}.
	
	However, implementing \gls{fl} over wireless networks raises several concerns, mainly due to the noisy nature of the wireless links connecting the end devices, as well as the limited computation and communication resources available at each client. 
	Along these lines, Yang \emph{et al.}, in~\cite{Yang2020}, tried to optimize both wireless and computational resources to minimize the learning training delay, even though considering the whole pool of clients at each round. 
	In turn, \gls{fl} methods typically select only a subset of devices at each iteration, in order to alleviate the burden of data transmission for distributing model updates. 
	For example, the authors in~\cite{MMAmiri2021} proposed a method to identify the optimal resource allocation policy as a function of the number of clients participating in the training process, while considering both channel conditions and the significance of their local model updates.
	The results suggest that the number of clients to be considered at each round should depend on how data are distributed on the local datasets: for \gls{iid} data, the best strategy is to sample just one client per round whereas, for a non-\gls{iid} scenario, the number of clients per iteration should be proportional to how heterogeneously the data are distributed, to avoid fitting locally skewed datasets.
	Another approach to reduce wireless resource occupancy during training is to compress and send sparse local model updates to the server, rather than quantizing the global model itself~\cite{MMAmiriQuantized2020}. 
	
	Despite these early results, however, it is still not clear, for non-\gls{iid} data distributions, how to quantify the trade-off between the number of clients involved in the training of the model and the number of training iterations that are needed to achieve a certain level of accuracy~\cite{BrendanMcMahan2017}.
	Moreover, most methods assume perfect knowledge of the \gls{csi} at each round, which is then leveraged to find the optimal power and resource allocation strategy to minimize the training time. However, perfect \gls{csi} may be difficult to obtain in practice, especially in case of fast fading channels.
	To the best of our knowledge, the only prior work attempting to analyze the \gls{fl} training process under channel uncertainty is~\cite{wadu2020federated}, where \gls{csi} is inferred using a \gls{gp} and radio resources are scheduled according to the estimated~\gls{csi}. However, the analysis does not investigate whether training times are affected by different channel statistics.
	
	Based on this introduction, in this work we propose a novel \gls{fl} training method, hereby referred to as Fixed Rate Federated Learning (FRFL), working under imperfect \gls{csi}\footnote{Unlike in other papers, where the expression \textit{imperfect CSI} denotes the presence of noise or errors in the channel estimation process, here we use it to mean that the only information available about the channel state is its statistical distribution.}, with frequency and time constraints. 
	More specifically, we analyze the convergence time as a function of different channel models (i.e., Rayleigh, Nakagami, and Rician, to characterize different fading regimes), data distributions (i.e., iid and non-iid), and the number of clients participating in the training of the model.
	Our contributions can be summarized as follows.
	\begin{itemize}
		\item We evaluate whether exploiting channel statistics, like the \gls{cdf} of the fading distribution, when perfect \gls{csi} is not available, can still help identify the optimal resource scheduling approach to minimize the convergence time. To do so, we investigate whether preventing clients that cannot sustain a minimum predefined transmission rate from sending model updates results in faster training.
		Numerical experiments show that, in Rayleigh channels, it is possible to reduce the convergence time by around $80\%$ with $90\%$ accuracy  if just half of the clients are able to successfully communicate, compared to a baseline in which all clients adopt the maximum achievable rate to transmit model data.
		
		
		\item We prove that, as expected, while admitting more clients at each round may not significantly affect the convergence time to achieve a certain accuracy, it can dramatically increase the probability of introducing stragglers into the loop. This effect is particularly remarkable in case of Rayleigh fading, compared to Nakagami and Rician, thus demonstrating how channel statistics should be considered as a bias to optimize scheduling policies for~\gls{fl}.
		
	\end{itemize}
	
	The rest of the paper is organized as follows. 
	In Sec.~\ref{sec:problem} we present our system model.
	In Sec.~\ref{sec:motivation} we describe the proposed FRFL method to reduce the convergence time in the training process in case of imperfect CSI.
	In Sec.~\ref{sec:analysis} we introduce our simulation settings and parameters, and discuss our numerical results. 
	Finally, Sec.~\ref{sec:conclusions_and_future_works} concludes the analysis with suggestions for future work.

	\section{System Model}
	\label{sec:problem}
	In FL, $N$ wireless devices cooperatively build a global model $g(\bm{\omega})$, stored into a central \gls{bs}, by sharing learning model updates derived from their local datasets $\mathcal{D}_n$, $n = 1, \dots, N$, which are a partition of the global dataset ${\mathcal{D} = \cup_n \mathcal{D}_n}$.
	The global model parameter vector is randomly initialized to $\bm{\omega}^0$. The training phase is then organized in rounds, indexed by $t$. 
	At the beginning of each round, the BS broadcasts the global parameters $\bm{\omega}^t$ to the clients. 
	Once received, each client $n$ can update its local model $g(\bm{\omega}_n)$, using a version of the \gls{sgd} algorithm~\cite{BrendanMcMahan2017}, by optimizing its local loss $F_n(g(\bm{\omega}_n^t), \mathcal{D}_n)$, which is a function of the local model $g(\bm{\omega}_n^t)$ and its dataset $\mathcal{D}_n$ at round $t$.
	At the end of the local optimization phase, the \gls{bs} selects a pool of $C^t$ clients, with $C^t \leq N$, to collaboratively upload their local model updates, which are then aggregated to generate a new global model that now exploits the knowledge acquired by the clients. The process continues until convergence.

	In this work we consider the situation in which the global model $g(\bm{\omega)}$ must be trained within a limited amount of time $T$, as described in~\cite{Buyukates2020}. For example, when a model is used to monitor/control a safety-critical process, e.g., in an \gls{iiot} scenario~\cite{Savazzi2021} or for teleoperated driving~\cite{zugno2020toward}, training data must be shared with low latency to guarantee that collaborative machines are synchronized.
	The problem can then be reformulated as follows.
	Assume that  $N$ wireless devices are connected to the \gls{bs} using wireless links in an \gls{ofdm} system. At each round, $C^t$ clients are selected and exclusively assigned an orthogonal channel of  bandwidth $B_k, \; k=1,\dots, K=C^t \leq N$~\cite{wadu2020federated,Chen2020TW}. From now on, we will refer to client $k$ as the one associated to the $k$-th channel.
	Communication links are modeled as slow fading channels. Unlike previous works, we consider the case in which the \gls{bs} does not have perfect \gls{csi}, but can estimate the \gls{cdf} $F(h)$ of the channel gain $h$.
	In principle, the maximum rate at which client $k$ can communicate its model parameters with arbitrarily low error probability at round $t$ is given by Shannon's formula
	\begin{equation}
		R_k^t = B_k \log_2\left(1 + h_k^t \frac{p_k^t \phi_k^t}{N_0 B_k}\right), \quad \forall k \in \{1, \dots, K\},
	\end{equation}  
	where $h_k^t$ is the channel gain, $p_k^t$ is the power allocated for transmission, and $\phi_k^t$ is the path loss experienced by client $k$ during iteration $t$, whereas $N_0$ is the noise power spectral density. 
	In our analysis, we consider the case in which clients adapt their power $p_k^t$ in such a way that the path loss and the noise are scaled to reach a constant and target quality factor $A$, which defines different SNR regimes,~i.e.,
	\begin{equation}
		A= \frac{p_k^t \phi_k^t}{N_0 B_k}, \quad \forall k \in \{1, \dots, K\}.
	\end{equation}
	In standard \emph{synchronous} \gls{fl}, the \gls{bs} has to wait until all $C^t$ clients involved in the training process at round $t$ upload their local updates before proceeding to the next round, thus the round duration depends on the time required by the slowest client to complete its local computations and update the model. 
	In this work, in turn, we will only consider \emph{communication heterogeneity} in FL~\cite{hosseinalipour2020federated}, and impose that each client performs its local computations within a constant time. 
	
	\section{Federated Learning Under Imperfect CSI: \\The Proposed Solution}
	\label{sec:motivation}
	As discussed in Sec.~\ref{sec:problem}, \gls{fl} methods typically consider a fixed number of clients $C^t$ to be involved in the training phase at round $t$, and then allocate radio resources in such a way that the time each client takes to upload its model updates within the round is minimized. 
	Different scheduling policies can be adopted depending on whether or not CSI is known a priori, as described in Sec.~\ref{ssec:perfect} and \ref{ssec:imperfect}, respectively.
	On one side, it is possible to reduce the number of rounds required for convergence by simply increasing $C^t$. For example, the analysis in~\cite{BrendanMcMahan2017} shows that, even under non-iid data distribution, increasing at each round the fraction of clients involved in the training process from $10\%$ to $100\%$ could halve the number of training rounds.
	On the other side, given synchronous \gls{fl}, the more clients participating at each round, the longer the time required to complete it.
	Indeed, we can trade the amount of information exchanged at each round, i.e., the client updates, with the total number of rounds that can be completed within a given time  $T$. Notably, if the \gls{bs} knows the \gls{cdf} of the channel fading distribution, it is possible to quantify how many model updates from the participating clients can be gathered in $T$ seconds. 
	
	
	\subsection{\Acrlong{sfl} (SFL) with Perfect CSI} 
	\label{ssec:perfect}
	In situations where fresh model updates must be distributed to the edge network with strict time constraints, the \gls{bs} should accept to complete one round even if some of the clients have yet not shared their federated data, thus increasing the overall number of rounds. 
	In a baseline \gls{sfl} approach with perfect channel knowledge, each client $k$ during round $t$ would select the optimal rate to communicate over the channel with arbitrarily low error probability as
	\begin{equation}
		R_k^t = B_k \log_2(1 + h_k^t A).
		\label{eq:r}	
	\end{equation}
	Let $Z$ be the size of the client's local vector parameter $\bm{\omega}_k$ (which is  equal to that of the global model $\bm{\omega}$), expressed in bits.
	The time required by client $k$ to reliably transmit the model updates in one round is then given by ${T_k^t = Z/R_k^t}$, which depends on the specific realization of the channel gain $h_k^t$, known a priori. With this consideration, we can see that, given the number of clients $C^t$ participating at round $t$, the round duration $T_{\rm round}^t$ is equal~to
	\begin{equation}
		T_{\rm round}^t = \max_{k =1, \dots, C^t} \left\{ \frac{Z}{R_k^t} \right\}. 
	\end{equation}
	When $C^t$ is large, $T_{\rm round}^t$ can rapidly grow out of control. Therefore, in practical \gls{sfl} applications, we shall set $T_{\rm round}^t \leq T_{\rm ths}$, so that $T_{\rm round}^t$ never exceeds a predefined threshold $T_{\rm ths}$.

	In this perspective, the rate that dominates the communication delay at round $t$ is determined by
	${h_{\rm m}^t = \min_k\{ h_k^t\}}_{k=1}^{C^t}$,
	whose \gls{cdf} and \gls{pdf} can be found, respectively, as
	\begin{equation}
		F_{\rm min}(h_{\rm m}^t) =  1-\Big[1-F(h_{\rm m}^t)\Big]^{C^t},
	\end{equation}
	\medmuskip=1mu
	\thickmuskip=1mu
	\begin{equation}
		f_{\rm min} (h_{\rm m}^t) = \frac{\partial F_{\rm min}(h_{\rm m}^t)}{\partial h_{\rm m}^t} = C^t\Big[1 - F(h_{\rm m}^t)\Big]^{(C^t-1)} f(h_{\rm m}^t),
	\end{equation}
	\medmuskip=6mu
	\thickmuskip=6mu
	
	where $F(h_{\rm m}^t)$ and $f(h_{\rm m}^t)$ are, respectively, the \gls{cdf} and the \gls{pdf} of the channel gain $h$ computed in $h_{\rm m}^t$.
	The round duration is therefore constrained by the minimum rate ${R_{\rm min}^t = \min_k \{R^t_k\}_{k=1}^{C^t}}$, i.e.,
	\begin{equation}
		T_{\rm round}^t = \frac{Z}{{R_{\rm min}^t}}=\frac{Z}{B_k \log_2(1 + h_{\rm m}^t A)}.
		\label{eq:T_csi}
	\end{equation}
	
	\subsection{\Acrlong{frfl} (FRFL) with Imperfect CSI} 
	\label{ssec:imperfect}
	In this section, we generalize the problem in Sec.~\ref{ssec:perfect} and assume that instantaneous channel information is not available at the server.
	If CSI is unknown, it is not possible to find the absolute optimal rate to minimize communication errors as in Eq.~\eqref{eq:r}. 
	We then propose a \gls{frfl} approach in which each client $k$ involved in the training process adopts a constant global rate ${R_k^t=R^*, \:\forall k \in \{1, \dots, K\}, \:\forall t}$, in such a way that it can complete each training round within ${T_{\rm round}^t  = T_{\rm round} = Z/R^* \leq T_{\rm ths}}$.
	From communication theory, it is well known that clients can communicate with rate $R^* \leq B_k \log_2(1 + h_k^t A )$ with arbitrarily low error probability.
	On the contrary, if the rate is such that ${R^* > B_k \log_2(1 + h_k^t A)}$, e.g., due to near-far effects or in a moving network, the packet error probability may rapidly grow to one, and the client participating in the training may not be able to communicate its model updates successfully.  This situation is also known as \emph{deep fading} condition~\cite{Angjelichinoski2019}. 
	In this case, the probability that the server loses the model updates sent from client $k$ at round $t$ is given by
	\begin{equation}
		\begin{split}
			\epsilon(R^*) & = \mathbb{P}[R^* > B_k \log_2(1 + h_k^t A)] \\[5pt]
			&= \mathbb{P}\left[ h_k^t < \left(\frac{2^{(R^*/B_k)} -1}{A}\right)\right] \\[5pt]
			&= F\left(\frac{2^{(R^* / B_k)} -1}{A}\right).
		\end{split}
	\end{equation}
	By exploiting the channel statistics, i.e., the \gls{cdf} $F(h)$ of the fading distribution $h$, the average number of clients $\hat{C}(R^*)$ successfully participating in each round $t$ when global rate $R^*$ is adopted can be quantified as
	\medmuskip=1mu
	\thickmuskip=1mu
	\begin{equation}
		\hat{C}(R^*) =C^t\Big[1-\epsilon(R^*)\Big]= C^t  \left[1-F\left(\frac{2^{(R^*/B_k)} -1}{A}\right)\right],
	\end{equation} 
	\medmuskip=6mu
	\thickmuskip=6mu
	where $C^t$ is the original pool of clients selected by the BS to communicate at round $t$.
	It appears clear that the choice of the optimal rate $R^*$ dominates the overall training performance.
	Indeed, $R^*$ can be adapted to include fewer or more clients in the training process, depending on the target number of iterations that must be completed within time $T$, and the average duration of each round. 
	In FRFL, we adopt a heuristic approach. The \gls{bs} first computes the expected minimum rate $\mathbb{E}[R_{\rm min}^t]$ experienced by the $C^t$ participating clients, and then selects $R^*$ such that $R^* >  \mathbb{E}[R_{\rm min}^t]$ if the corresponding error $\epsilon(R^*)$ is below an arbitrary threshold that is deemed acceptably low to allow proper accuracy in the~training.
	By the convexity of the function $\psi(R_{min}^t) = 1/R_{\rm min}^t$ and Jensen's inequality, it results that $\mathbb{E}[T_{\rm round}^t] \geq Z/\mathbb{E}[R_{\rm min}^t]$: using a fixed rate $R^* = \alpha \mathbb{E}[R_{\rm min}^t]$, with $\alpha > 1$, results in a reduction of the lower bound for the average round duration compared to the baseline SFL method, as expressed in Eq.~\eqref{eq:T_csi}, by a factor $\alpha$, as we will demonstrate in Sec.~\ref{sub:numerical_results}.
	We do not preclude more sophisticated methods, e.g., based on mathematical analyses or reinforcement learning, to be adopted for selecting $R^*$, even though this is out of the scope of this paper and will be part of our future~work.

	\section{Performance Results}
	\label{sec:analysis}
	
	In this section, we describe our simulation settings, i.e., the channel models (Sec.~\ref{sub:channel}) and parameters (Sec.~\ref{sub:simulation_parameters}) we adopt, and present our numerical results (Sec.~\ref{sub:numerical_results}).
	
	\subsection{Channel Models}
	\label{sub:channel}
	Unlike most literature analyses, in this work we  characterize the FL training performance as a function of different channel models, so as to incorporate the effect of different fading regimes.\footnote{Notice that, while Rayleigh fading is generally assumed for transmissions in the legacy bands, 5G and beyond communication systems may operate in new spectrum bands, e.g., the lower part of the millimeter wave (mmWave) bands~\cite{rappaport2013millimeter}, where a Rician or Nakagami model would better characterize the effect of multi path components, as expected at those frequencies~\cite{lecci2020simplified}.}
	Let $F(h)$ be the \gls{cdf} of the channel gain $h$, where in the rest of the analysis we omit indices $k$ and $t$ to indicate the client and the round, respectively, under the assumption that channel realizations are \gls{iid} in frequency and time. The following channel models are considered~\cite{Angjelichinoski2019}.
	
	\paragraph{Rayleigh channel}
	The Rayleigh channel model represents a single diffuse component~\cite{Durgin2002}, and is one of the most widely adopted channel models in wireless communications thanks to its simplicity and mathematical tractability. Let $\sigma^2$ denote the average squared channel gain, i.e., $\mathbb{E}[h^2] = \sigma^2$; the \gls{cdf} $F(h)$ of $h$ is then computed as
	\begin{equation}
		F (h) = 1 - e^{-\frac{h^2}{2\sigma^2}}, \; h \geq 0
	\end{equation}   
	In our experiment, we consider the standard Rayleigh parameterization with $\sigma^2 = 1$, as typically considered in legacy communication systems.

	\paragraph{Rician channel} 
	The Rician distribution is usually adopted to model an additional dominant, specular, multi path component from the transmitter to the receiver~\cite{Durgin2002}. The channel is parameterized by the factor $K = \nu^2/(2\sigma^2)$, where $\nu^2$ is the contribution of the multi path component power, and $\sigma^2$ is related to the diffuse component, as in the Rayleigh case. The \gls{cdf} $F(h)$ of $h$ is given by
	\begin{equation}
		F(h) = 1 - Q_1\left(\frac{\nu}{\sigma}, \frac{h}{\sigma}\right), \; h \geq 0
	\end{equation}
	where $Q_1$ is the Marcum Q-function. We parameterize the Rician model with $K=12$ dB~\cite{Samimi2016}.
	
	\paragraph{Nakagami channel} 
	The Nakagami distribution extends the Rayleigh model to incorporate multiple clusters, and is parameterized by the shape parameter $m$, which represents the number of \gls{iid} diffuse components, each modeled as a Rayleigh distribution with mean diffuse power $\sigma^2$~\cite{NAKAGAMI19603}. The corresponding \gls{cdf} $F(h)$ of $h$ is given by
	\begin{equation}
		F(h) = \frac{\gamma(m, \frac{m}{\sigma^2} h^2)}{\Gamma(m)}, \; h \geq 0 
	\end{equation}
	where $\gamma(\cdot, \cdot)$ is the lower incomplete Gamma function, and $\Gamma(\cdot)$ is the Gamma function.
	In this paper we set $m=3$~\cite{heath2015}.

	\begin{figure}[t!]
		\centering
		\includegraphics[width=0.95\columnwidth]{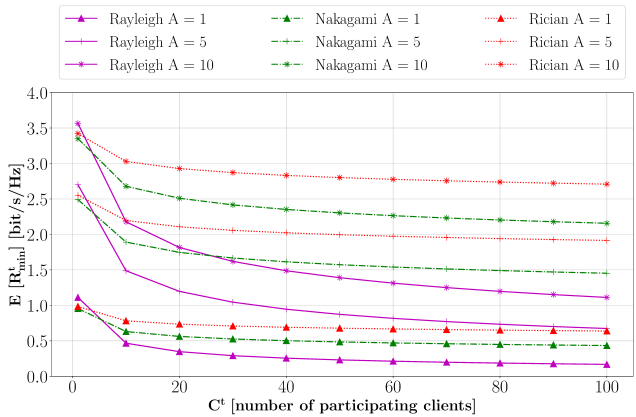}  
		\caption{Average minimum rate/Hz vs. number of participating clients for the SFL policy in case of Rayleigh, Nakagami and Rician channels and for different values of the quality factor $A$. }
		\label{fig:minimum_avg}
	\end{figure}
	
	In Fig.~\ref{fig:minimum_avg} we plot the average minimum rate $\mathbb{E}[R_{\rm min}^t]$ for different channel distributions, as a function of the  quality factor $A$ and the number of clients $C^t$ that participate in a generic training round $t$, when perfect CSI is available. 
	We observe that $\mathbb{E}[R_{\rm min}^t]$ decreases significantly as the number of clients increases, especially when Rayleigh channels are considered. This is expected as the Nakagami and Rician models present a smaller variance. For example, in the presence of poor Rayleigh channel conditions, e.g., $A = 1$, the average minimum rate drops by more than $50\%$, resulting in more than twice the training delay, when only $10$ clients are involved in each round. The same effect is observed even in case of strong channels, i.e., $A = 10$, and if $40$ clients selected to~participate.

	\begin{figure*}[t!]
		\centering
		\begin{subfigure}{0.32\textwidth}
			\centering
			\includegraphics[width=1\linewidth]{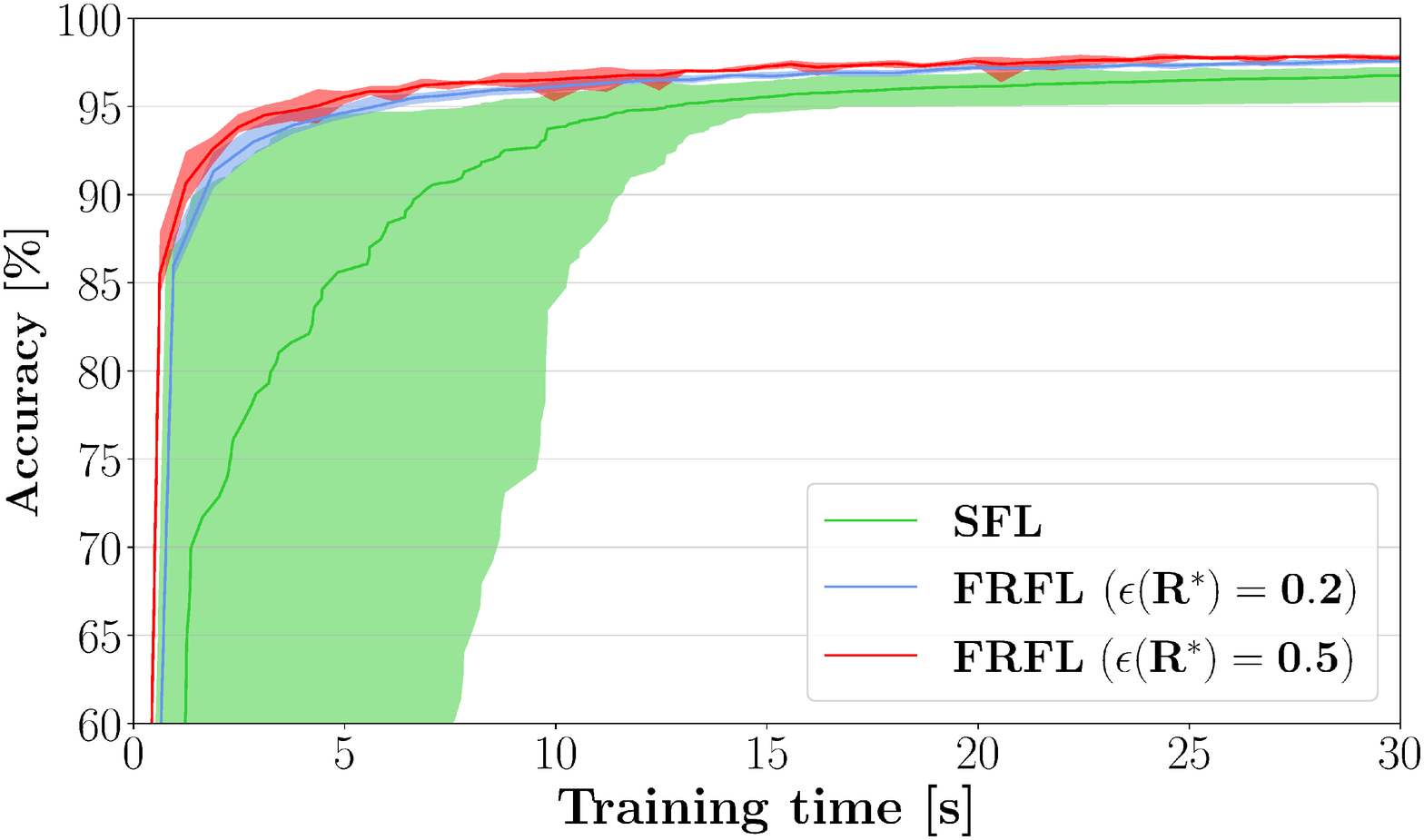}  
			\caption{$C^t= 10$ and $A=1$.}
			\label{fig:10-1}
		\end{subfigure}
		\begin{subfigure}{0.32\textwidth}
			\centering
			\includegraphics[width=1\linewidth]{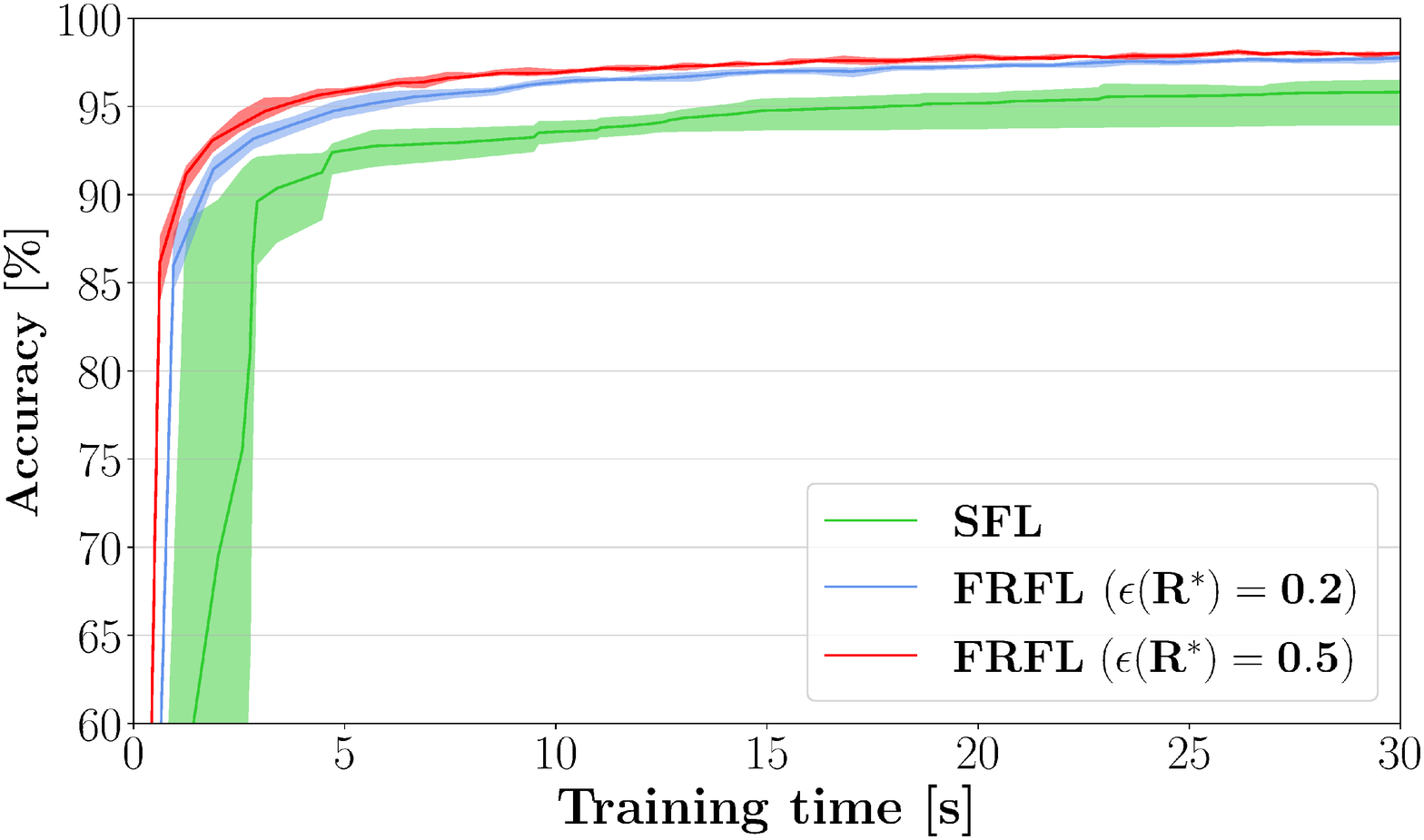}  
			\caption{$C^t= 20$ and $A=1$.}
			\label{fig:20-1}
		\end{subfigure}
		\begin{subfigure}{0.32\textwidth}
			\centering
			\includegraphics[width=1\linewidth]{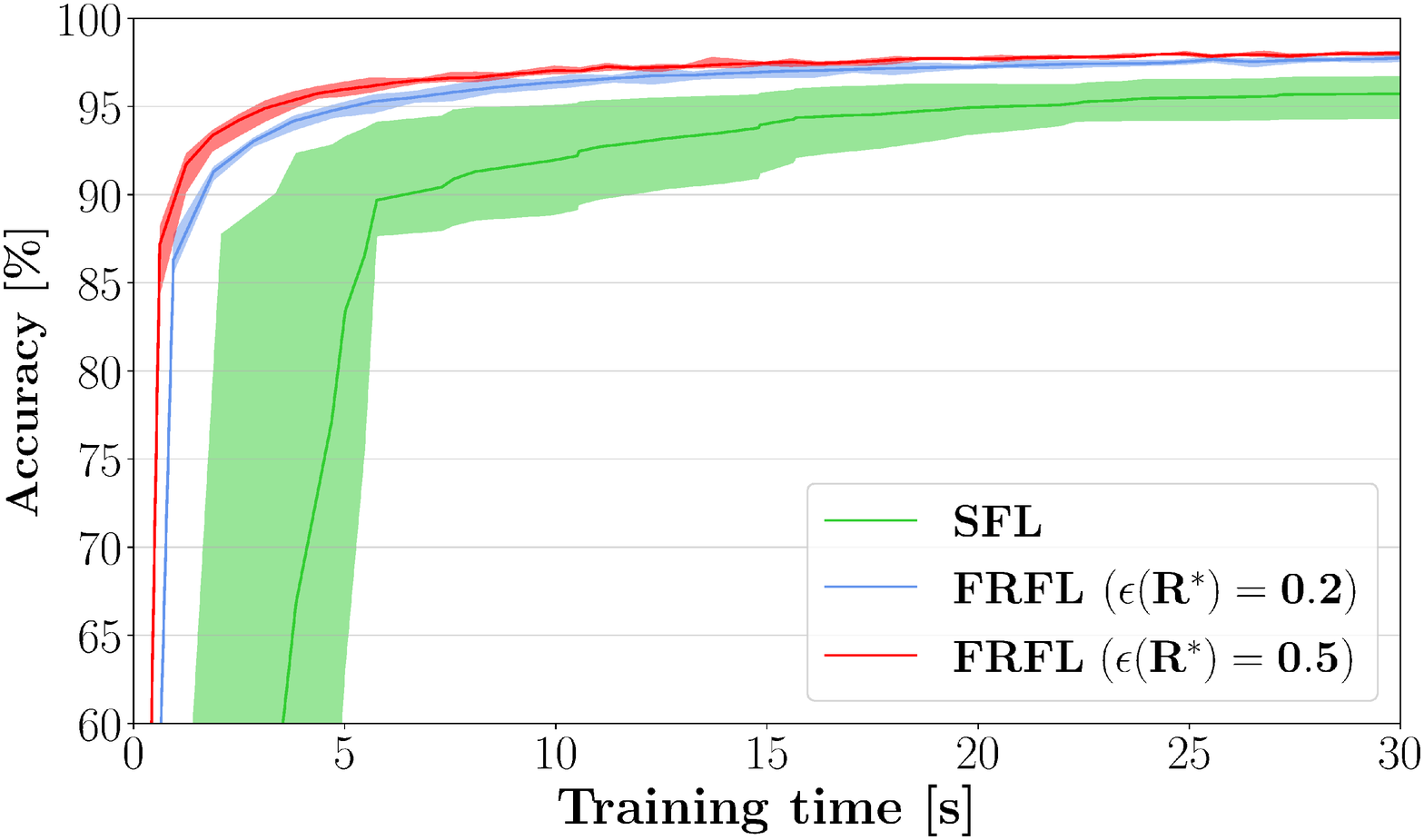}  
			\caption{$C^t= 40$ and $A=1$.}
			\label{fig:40-1}
		\end{subfigure}
		\begin{subfigure}{0.32\textwidth}
			\centering
			\includegraphics[width=1\linewidth]{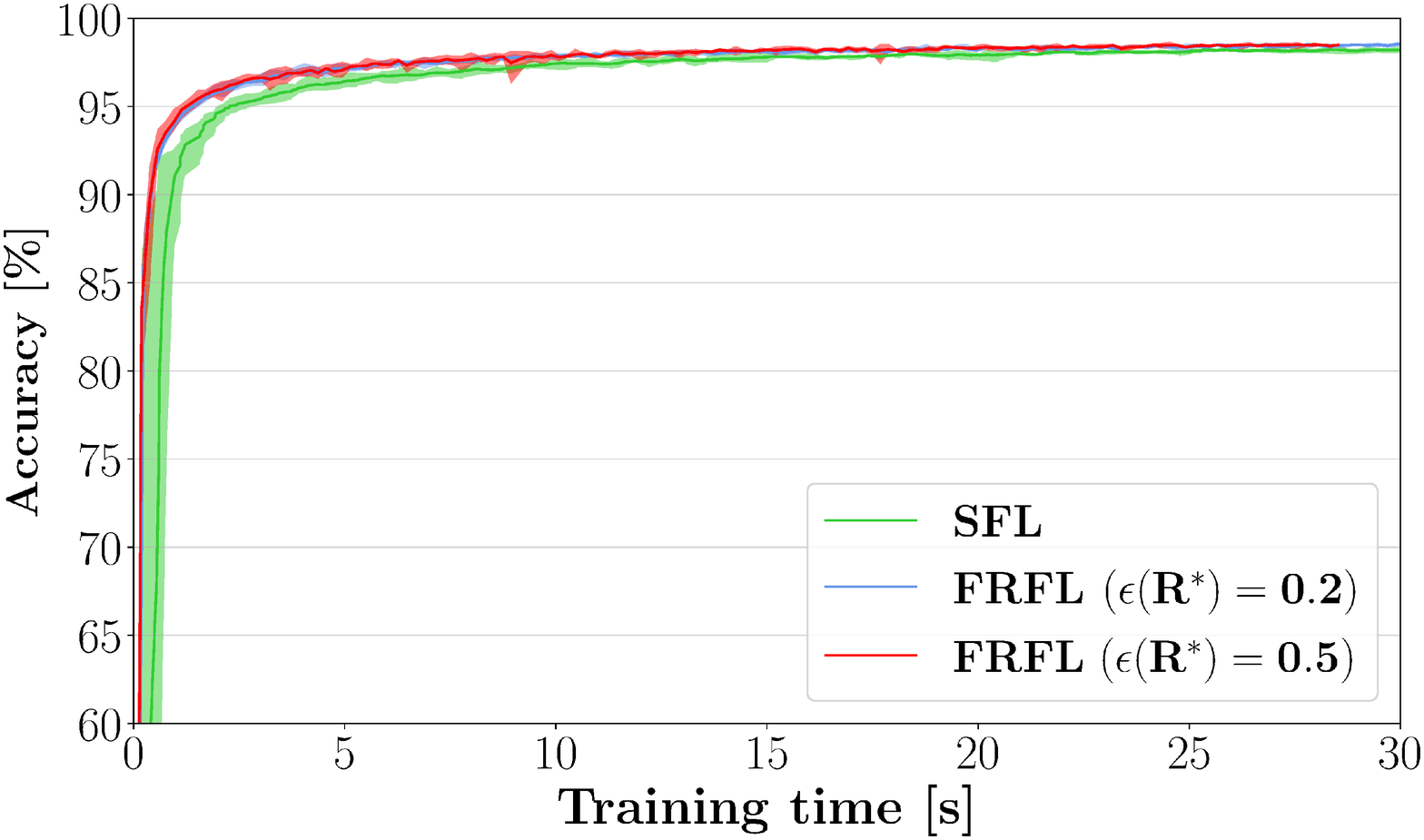}  
			\caption{$C^t=10$ and $A=10$.}
			\label{fig:10-10}
		\end{subfigure}
		\begin{subfigure}{0.32\textwidth}
			\centering
			\includegraphics[width=1\linewidth]{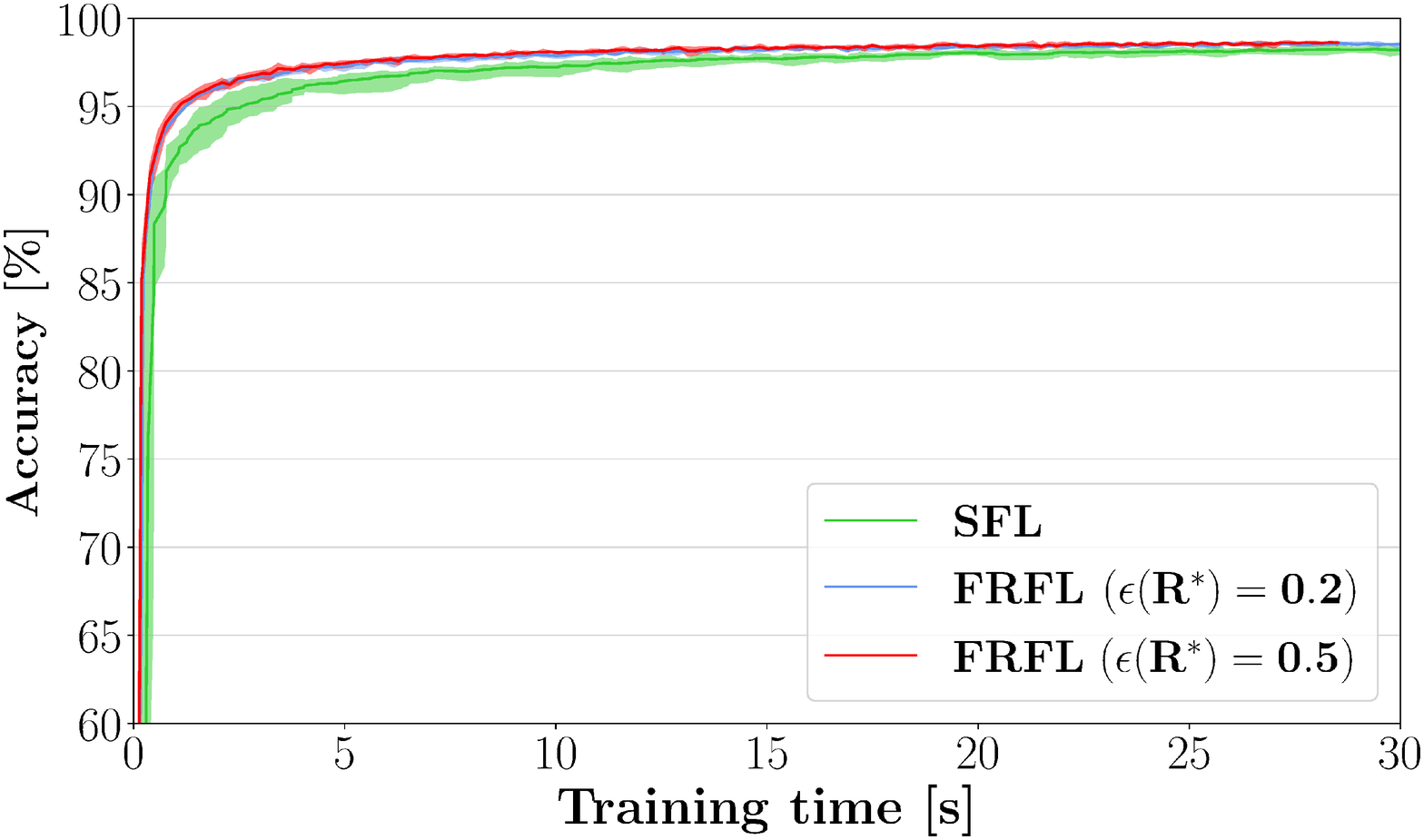}  
			\caption{$C^t= 20$ and $A=10$.}
			\label{fig:20-10}
		\end{subfigure}
		\begin{subfigure}{0.32\textwidth}
			\centering
			\includegraphics[width=1\linewidth]{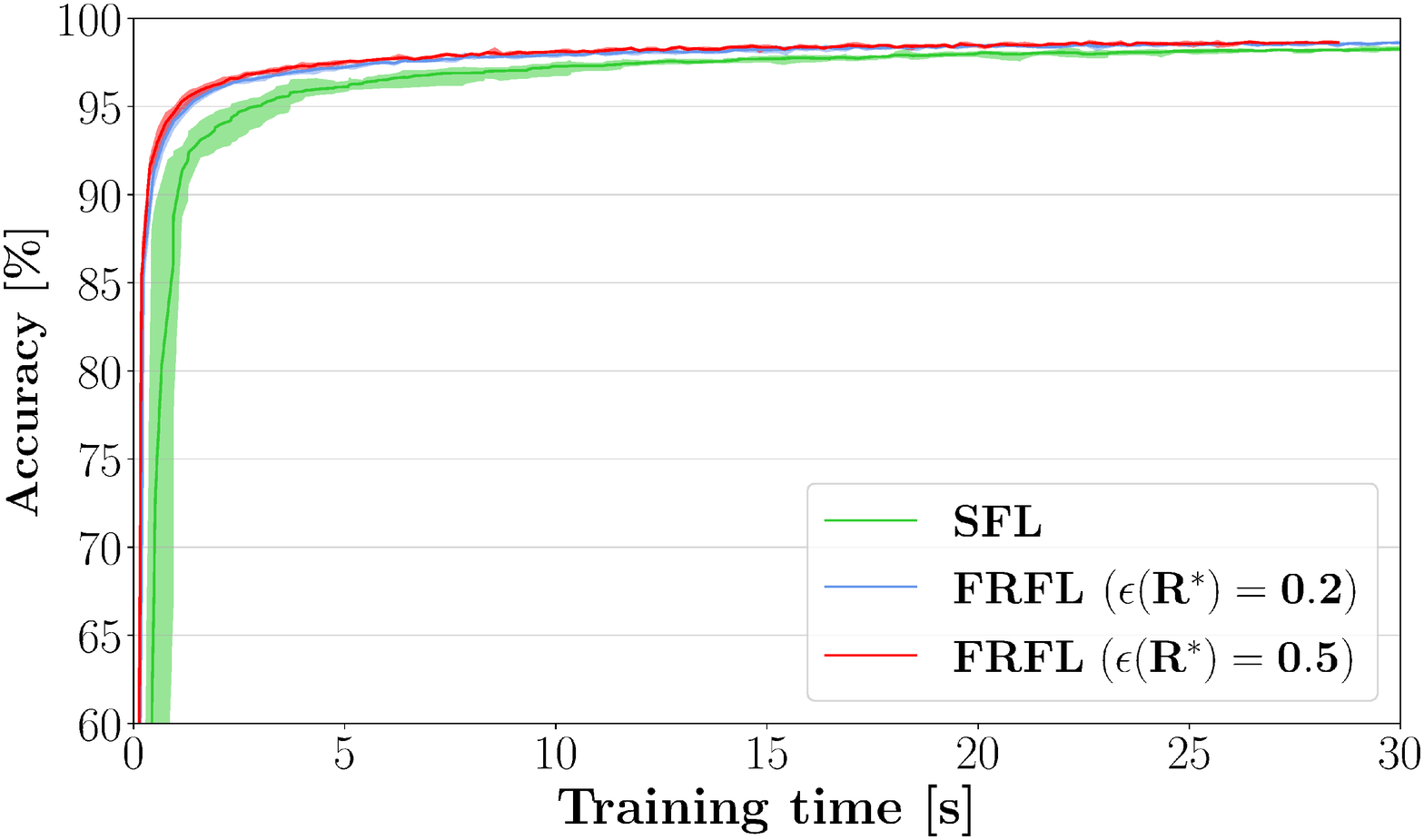}  
			\caption{$C^t= 40$ and $A=10$.}
			\label{fig:40-10}
		\end{subfigure}
		\caption{Min-to-max and average accuracy (over $5$ simulations) during the training process as a function of the time and the number of clients $C^t$ involved in the rounds, considering both SFL and FRFL methods. Rayleigh fading with $A=1$ (first row) and $A=10$ (second row), and \gls{iid} data are considered.}
		\label{fig:iid_rayleigh_curves}
	\end{figure*}

	\begin{figure*}[t!]
		\centering
		\begin{subfigure}{0.32\textwidth}
			\centering
			\includegraphics[width=1\linewidth]{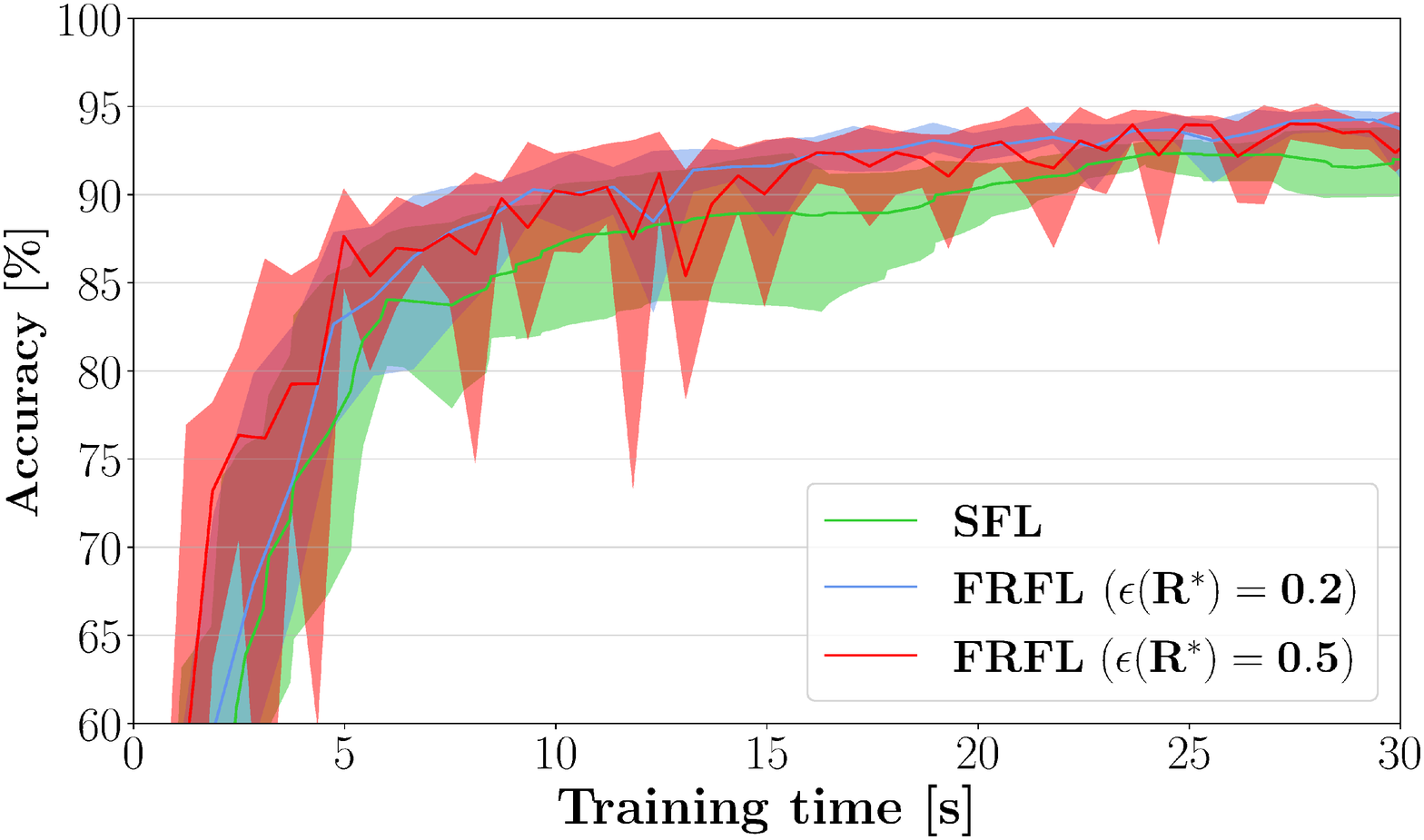} 
			\caption{$C^t=10$ and $A=1$.}
			\label{fig:rayleigh_curve_non_iid_10}
		\end{subfigure}
		\begin{subfigure}{0.32\textwidth}
			\centering
			\includegraphics[width=1\linewidth]{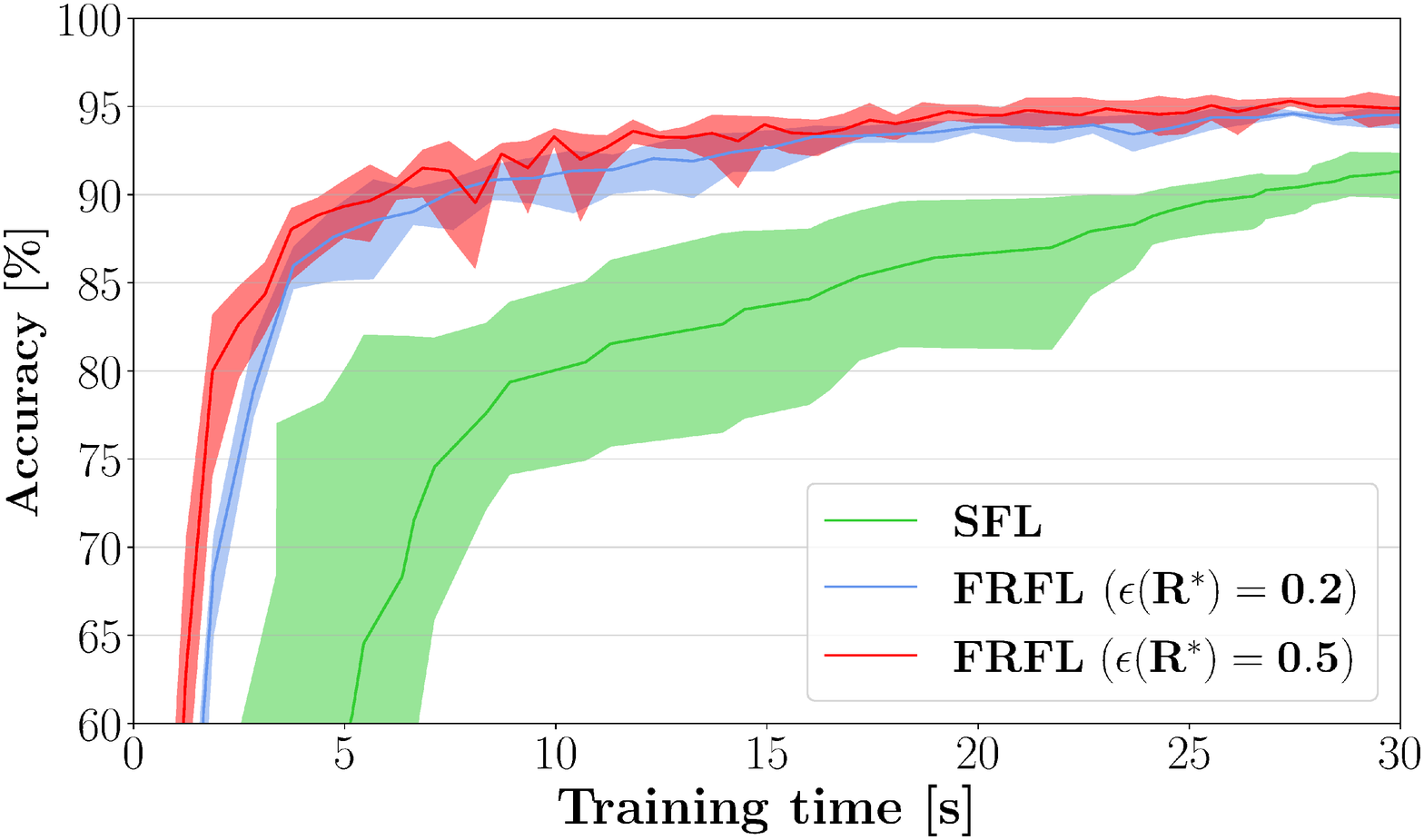}  
			\caption{$C^t=20$ and $A=1$.}
			\label{fig:20-1-non}
		\end{subfigure}
		\begin{subfigure}{0.32\textwidth}
			\centering
			\includegraphics[width=1\linewidth]{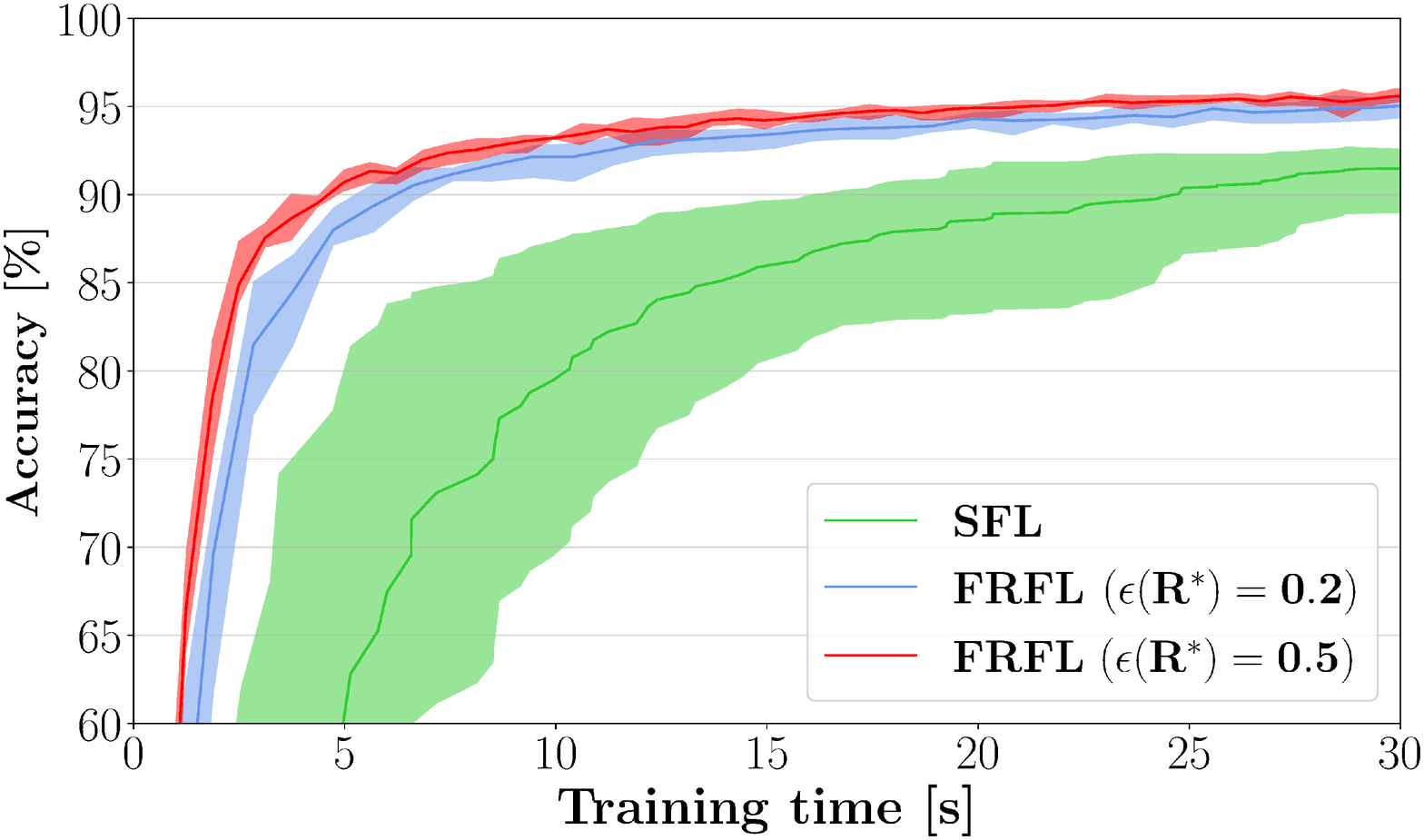}  
			\caption{$C^t=40$ and $A=1$.}
			\label{fig:40-1-non}
		\end{subfigure}
		\begin{subfigure}{0.32\textwidth}
			\centering
			\includegraphics[width=1\linewidth]{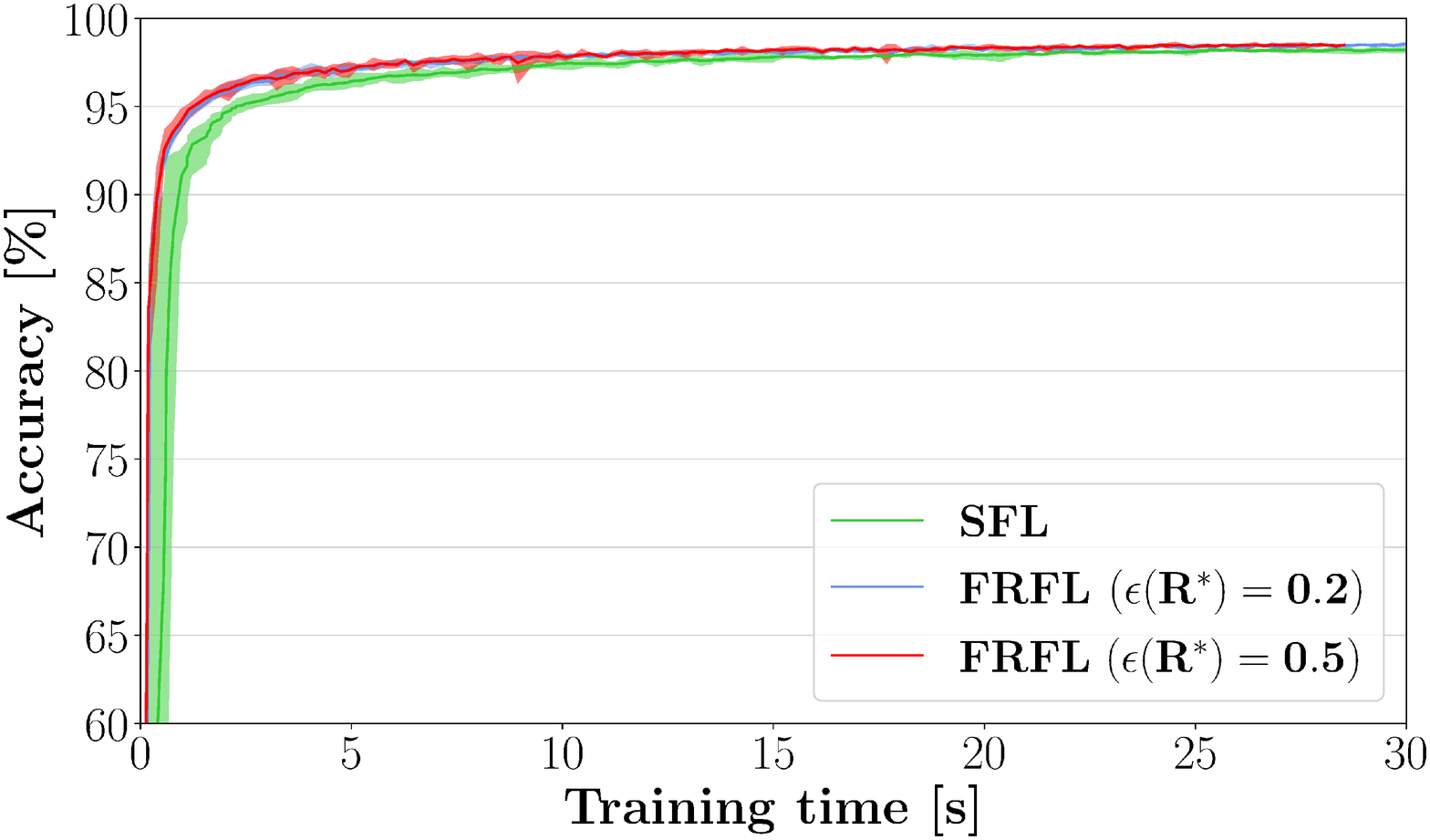}  
			\caption{$C^t=10$ and $A=10$.}
			\label{fig:}
		\end{subfigure}
		\begin{subfigure}{0.32\textwidth}
			\centering
			\includegraphics[width=1\linewidth]{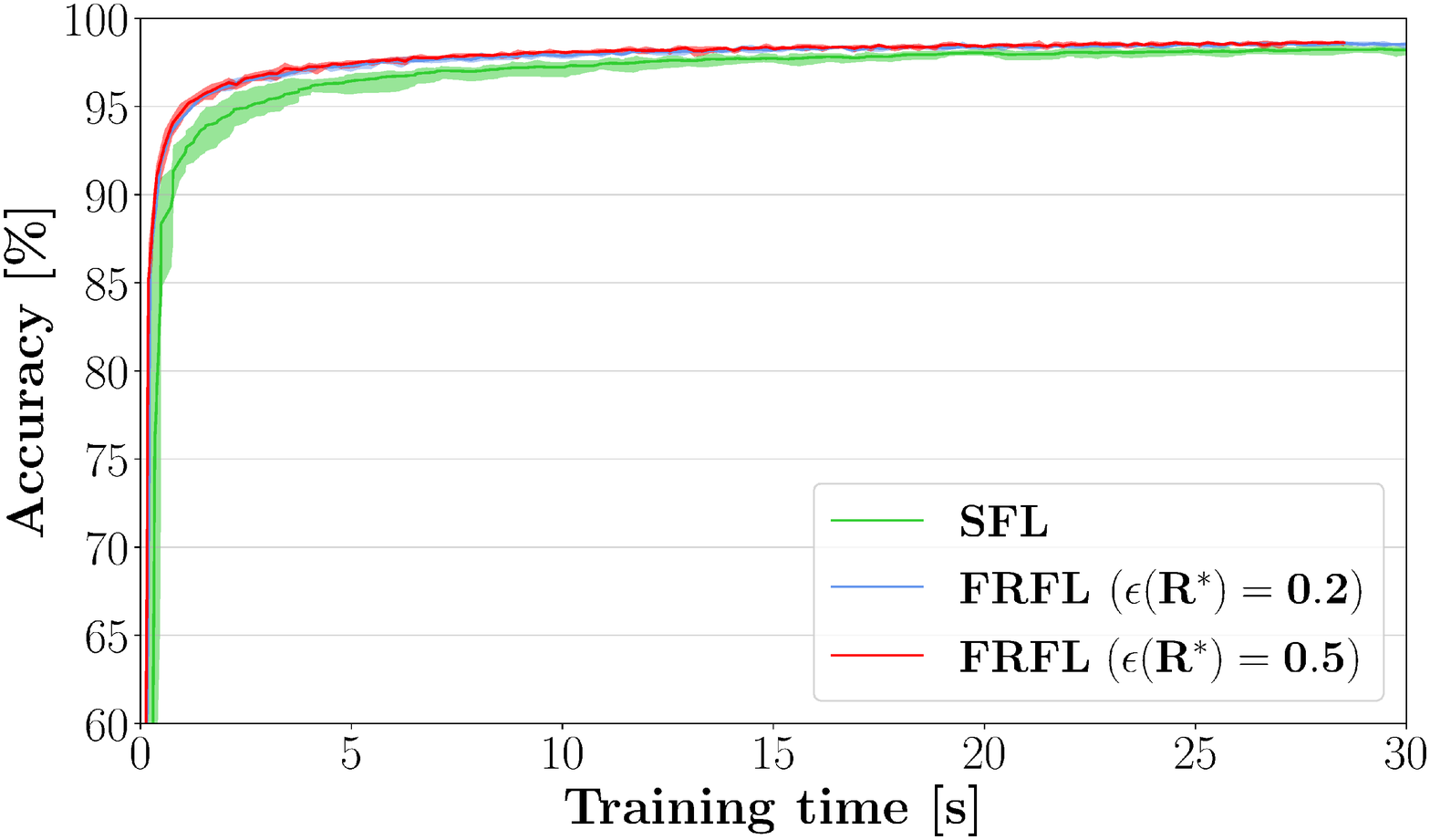}  
			\caption{$C^t=20$ and $A=10$.}
			\label{fig:}
		\end{subfigure}
		\begin{subfigure}{0.32\textwidth}
			\centering
			\includegraphics[width=1\linewidth]{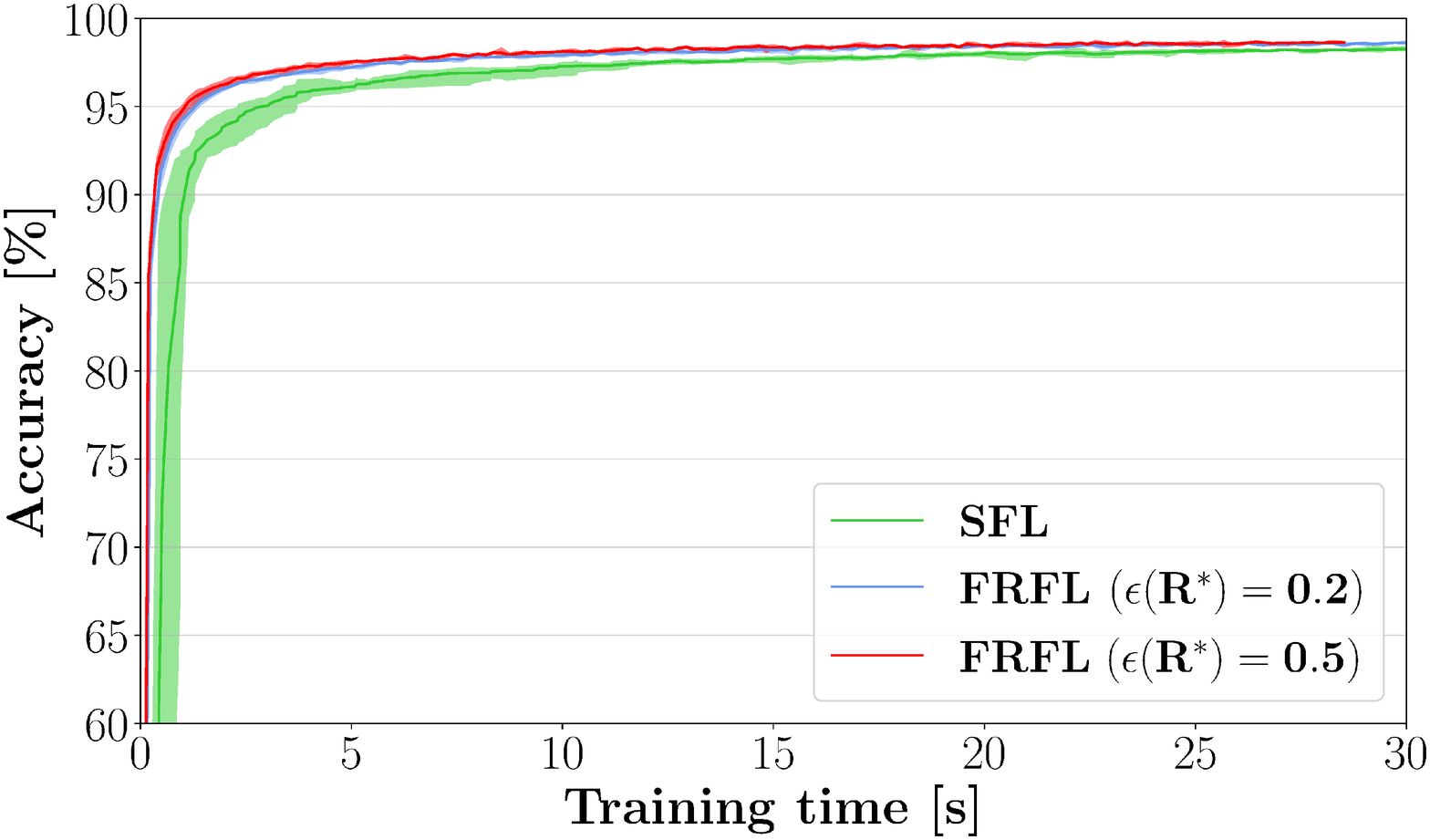}  
			\caption{$C^t=40$ and $A=10$.}
			\label{fig:}
		\end{subfigure}
		\caption{Min-to-max and average accuracy (over $5$ simulations) during the training process as a function of the time and the number of clients $C^t$ involved in the rounds, considering both SFL and FRFL methods. Rayleigh fading with $A=1$ (first row) and $A=10$ (second row), and non-\gls{iid} data are considered.\vspace{-0.18cm}}
		\label{fig:non_iid_rayleigh_curves}
	\end{figure*}

	\subsection{Simulation Parameters and Setting} 
	\label{sub:simulation_parameters}
	
	Based on the results in Fig.~\ref{fig:minimum_avg}, in our simulations we consider $N=100$ overall wireless clients, while only $C^t \in \{10,\,20,\,40\}$ of them are selected to participate in the model updates at generic round $t$. 
	Each participating client uses an orthogonal channel of $1$ MHz of bandwidth in all the investigated configurations. Two different values of $A$, i.e., $1$ and $10$, are considered in the Rayleigh case, with iid and non-iid data distributions, whereas $A=1$ is selected for Rician and Nakagami channels. In our experiments we evaluate the performance of the FL training process, specifically the convergence time, comparing two different scheduling strategies: 
	a baseline SFL approach with full channel information (Sec.~\ref{ssec:perfect}), and two different versions of the FRFL strategy working under imperfect CSI (Sec.~\ref{ssec:imperfect}), with $\epsilon(R^*)=0.2$ and $0.5$. The two models assume that on average $20\%$ and $50\%$ of the clients, respectively, are not able to communicate their training updates due to bad channel conditions at the selected global rate $R^*$. In both cases, $\epsilon(R^*)$ has been selected so that $R^*>\mathbb{E}[R_{\rm min}^t]$ in all simulation scenarios. 
	The training time is set to $T = 30$ seconds,  which is large enough to let the model be trained with an acceptable level of accuracy.

	The simulations are conducted on the MNIST dataset \cite{lecun-mnist}, which contains $70\,000$ ($60\,000$ for training and $10\,000$ for testing)  handwritten digits, classified into one of $10$ possible classes. 
	While, for \gls{iid} data distribution, each client has $600$ training samples, and classes are uniformly distributed among the local datasets, in the non-iid setting a random number of training samples and classes are distributed among the clients.

	The learning model is a \gls{cnn} with two $5 \times 5$ convolutional layers (with $10$ and $20$ channels and a $2 \times 2$ max pooling operation after the first layer), followed by one dense layer with $320$ neurons and one output layer with $10$ units. 
	The activation function for the inner layers is the ReLu function, whereas softmax is used for the output layer. The loss is modeled by the cross-entropy function, which is a standard option in classification problems.
	Training weights are aggregated at the \gls{bs} according to the {FedAvg} aggregator function~\cite{BrendanMcMahan2017}: at the end of round~$t$, the new global vector parameter $\bm{\omega}^{t+1}$ is computed as 
	\begin{equation}
		\bm{\omega}^{t+1} = \frac{1}{D^t} \sum_{k=1}^{C^t} D_k^t \bm{\omega}_k^t,	
	\end{equation}
	where $D^t = \sum_{k=1}^{C^t} D_k^t$, with $D_k^t$ being the size of the local dataset $\mathcal{D}_k^t$, and the local parameter vectors $\{ \bm{\omega}_k^t\}_{k=1}^{C^t}$ are updated using the \gls{sgd} algorithm with momentum equal to $0.5$ and learning rate set to $0.01$. 
	Notice that, in FRFL, some clients may not be able to share their local parameter vectors. Therefore, if client $k$ experiences a transmission error during round $t$, $ \bm{\omega}_k^t$ is set to $\bm{0}$ at the \gls{bs}, and $D_k^t=0$.
	

	\subsection{Numerical Results}
	\label{sub:numerical_results}

	\begin{figure}[t!]
		\centering
		\begin{subfigure}{0.45\textwidth}
			\centering
			\includegraphics[width=0.9\linewidth]{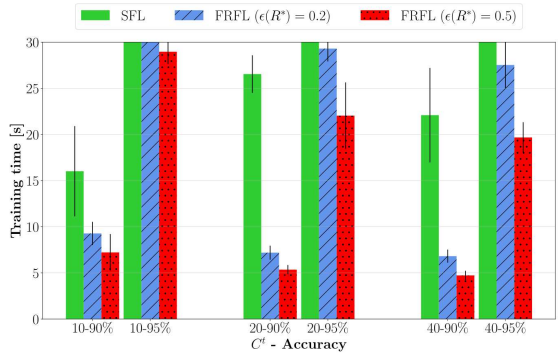}
			\caption{Chanel quality factor $A=1$.}
			\label{fig:rayleigh_accuracy_1}
		\end{subfigure}\\[5pt]
		\begin{subfigure}{0.45\textwidth}
			\centering
			\includegraphics[width=0.9\linewidth]{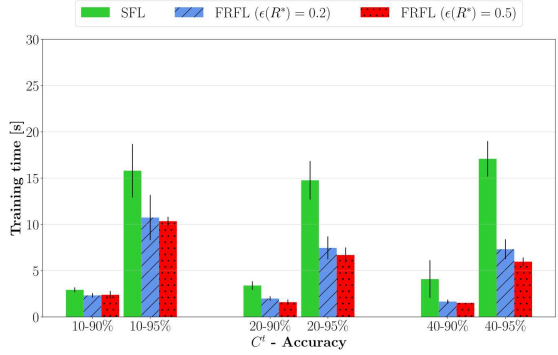}  
			\caption{Chanel quality factor $A=10$.}
			\label{fig:rayleigh_accuracy_10}
		\end{subfigure}
		\caption{Average time (and confidence intervals) to achieve $90\%$ and $95\%$ accuracy in Rayleigh fading channels, as a function of the number of clients $C^t$ involved in the rounds, considering both SFL and FRFL methods.\vspace{-0.33cm}}
		\label{fig:rayleigh_accuracy}
	\end{figure}
	
	In this section we validate the performance of the proposed FRFL method when imperfect CSI is considered.
	Fig. \ref{fig:iid_rayleigh_curves} plots the average accuracy over time achieved on the test dataset during the federated training process in Rayleigh channels, as a function of the number of clients $C^t$ involved in the training and the channel condition $A$, and assuming \gls{iid} data.
	
	First, we observe that adding more clients per round does not impact the long-term accuracy even with imperfect CSI, as acknowledged by prior analyses, e.g., in~\cite{MMAmiri2021}. In fact, FRFL assumes a fixed global rate $R^*$ for all participating clients, which does not affect the transmission delay.
	On the contrary, in case CSI is available, SFL implies that the more clients involved in the communications rounds, the longer, on average, the time it takes for the server to receive all model updates, which results in slower convergence.
	For example, at 5 seconds, the accuracy drops from around 95\% to 85\% when SFL is considered, for $C^t =10$ and $A=1$.
	Fig. \ref{fig:10-1}, Fig. \ref{fig:20-1}, and Fig. \ref{fig:40-1} further demonstrate that considering a weaker channel, i.e., $A=1$, degrades the long–term accuracy performance of the training, as adding more clients slows down the communications rounds.
	In case of more robust channels with $A=10$ (Fig. \ref{fig:10-10}, Fig. \ref{fig:20-10}, and Fig. \ref{fig:40-10}) this effect is mitigated, e.g., at 5 seconds, for  $C^t =10$, the SFL training accuracy increases by around 13\% compared to $A=1$. In any case, FRFL always outperforms SFL, even in the presence of perfect CSI.

	\begin{figure}[t!]
		\centering
		\begin{subfigure}{0.45\textwidth}
			\centering
			\includegraphics[width=0.9\linewidth]{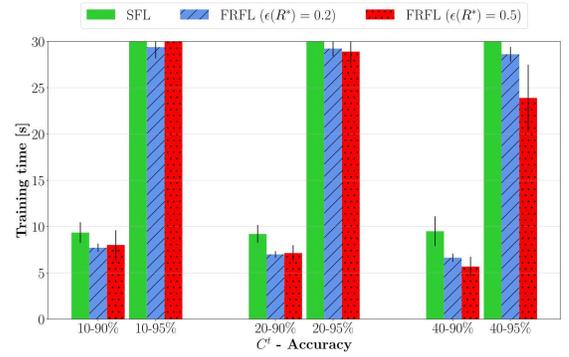}
			\caption{Nakagami channel and $A = 1$.}
			\label{fig:nakagami_accuracy}
		\end{subfigure}\\[5pt]
		\begin{subfigure}{0.45\textwidth}
			\centering
			\includegraphics[width=0.9\linewidth]{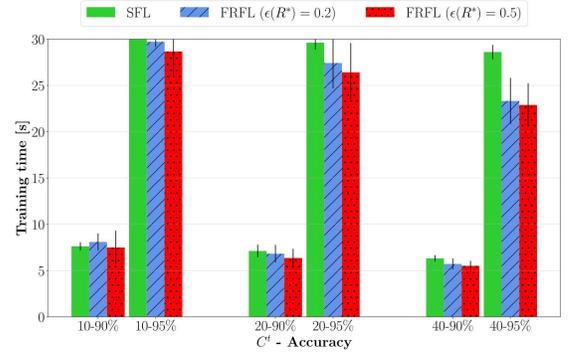}  
			\caption{Rician channel and $A = 1$.}
			\label{fig:rician_accuracy}
		\end{subfigure}
		\caption{Average time (and confidence intervals) to achieve $90\%$ and $95\%$ accuracy in Nakagami and Rician fading channels, as a function of the number of clients $C^t$ involved in the rounds, considering both SFL and FRFL~methods.\vspace{-0.33cm}}
		\label{fig:others_accuracy}
	\end{figure}
	
	In Fig. \ref{fig:non_iid_rayleigh_curves}, the SFL vs. FRFL performance is evaluated with non-\gls{iid} data. 
	In this case, gathering information from a smaller fraction of clients cannot generally sustain sufficiently high levels of accuracy. For example, Fig.~\ref{fig:rayleigh_curve_non_iid_10} presents an accuracy always lower than 95\% for $C^t=10$ in all investigated configurations.
	Increasing the number of clients may improve the accuracy performance during the whole training time, even though this effect is mitigated in the SFL strategy as the more participating clients imply also longer round durations.
	Moreover, it is interesting to compare the results for the SFL policy with $C^t=20$ (Fig.~\ref{fig:20-1-non}) and the FRFL policy with $C^t=40$ and $\epsilon(R^*) = 0.5$ (Fig.~\ref{fig:40-1-non}). 
	In both cases, the training involves 20 participating clients, as FRFL implies that, on average, $50\%$ of the clients do not successfully deliver their model updates on time, i.e., $\hat{C}(R^*)=20$. Then, even though the FRFL approach achieves better accuracy than SFL despite imperfect CSI (i.e., $95\%$ vs. $90\%$ at the end of the training when $A=1$), it requires $40$ channels to be allocated to the  $C^t=40$ clients, thus consuming twice the frequency resources. However, better performance against SFL can still be guaranteed with $C^t = 20$, that in turn requires $20$ orthogonal channels for both policies.

	Fig. \ref{fig:rayleigh_accuracy} compares the training time required to obtain $90\%$ and $95\%$ accuracy in Rayleigh channels with $A=1$ and $A=10$, when either SFL or FRFL is considered, as a function of the number of clients involved in the rounds. 
	First, we observe that it is possible to converge faster by trading the amount of information collected at each round with the round duration, which in turn increases the total number of possible rounds within $T=30$ s.  
	For example, Fig. \ref{fig:rayleigh_accuracy_1} shows that, when $A=1$ and $C^t=20$, the training time to reach $90\%$ accuracy can be reduced by almost $80\%$ if the proposed FRFL training method is adopted. Moreover, when $A=1$, the baseline SFL configuration, which always tends to assign the largest possible rate to its participating clients, is never able to reach $95\%$ accuracy within the training time despite leveraging full CSI. In turn, the FRFL policy with $\epsilon(R^*) = 0.5$ and $C^t=40$ succeeds in only $20$ seconds, on average, with small deviations. 
	\begin{figure}[t!]
		\setlength{\belowcaptionskip}{-0.33cm}
		\centering
		\includegraphics[width=0.95\columnwidth]{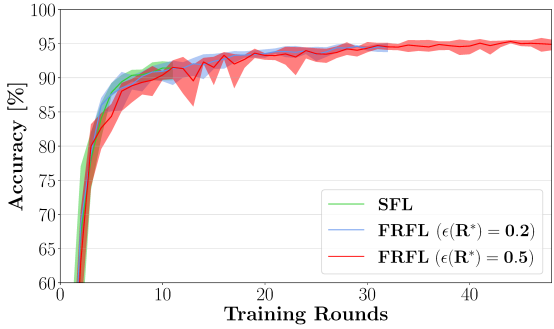}
		\caption{Min-to-max and average accuracy (over $5$ simulations) during the training process as a function of the number of rounds with Rayleigh fading ($A=1$), $C^t=20$, and non-\gls{iid} data. }
		\label{fig:number_of_rounds}
	\end{figure}
	The same conclusions can be derived from Fig.~\ref{fig:others_accuracy}, which investigates the impact of different channel models, i.e., Nakagami (Fig.~\ref{fig:nakagami_accuracy}) and Rician (Fig.~\ref{fig:rician_accuracy}), on the convergence time, for $A=1$. 
	First, we notice that, even though Rayleigh channels guarantee, on average, higher gains in single-link communications, Nakagami and Rician channels can support faster convergence for both SFL and FRFL policies: with Rician fading, for $C^t=10$, SFL with perfect CSI obtains $95\%$ accuracy in less than 10 seconds, against the 16 seconds when Rayleigh is adopted. This can be explained by the fact that both Nakagami and Rician fading exhibit lower variance, and can admit more clients per round, without increasing the average delay considerably.
	Nevertheless, the proposed FRFL policy always achieves faster convergence even with imperfect \gls{csi} by configuring faster rounds. Finally, Fig.~\ref{fig:number_of_rounds} depicts the training accuracy as a function of the number of rounds, in case of Rayleigh fading with $A=1$, $C^t=20$, and non-iid data. It is possible to see that, within the allocated time $T=30$ s, the FRFL policy with $\epsilon(R^*)=0.2$ ($\epsilon(R^*)=0.5$) is able to operate though 32 (48) rounds, while in turn the SFL policy is limited to 12 rounds, and never achieves 95\% accuracy. As a consequence, our analysis demonstrates that it may be convenient to neglect model updates from some participating clients, e.g., the most channel-constrained devices, as per the FRFL strategy, in favor of more round opportunities during training.
	The same trend is illustrated in Fig. \ref{fig:rayleigh_accuracy_1} with $C^t=20$.

	\section{Conclusions and Future Works} 
	\label{sec:conclusions_and_future_works}
	Federated learning is emerging as one the most popular distributed learning algorithms in which wireless devices collaboratively learn a global model without sharing training data.
	In this work we propose a novel federated learning method that decreases the convergence time by assigning a global constant rate to all the clients participating in the training rounds. Notably, the proposed approach does not require CSI availability, unlike most existing analyses.
	Our simulation results, validated in different channel regimes, demonstrate that the proposed approach, despite   considering imperfect CSI, always achieves better training performance compared to a baseline strategy in which the clients always adopt the maximum achievable rate to transmit model data. As part of our future work, we will investigate how to select the global constant rate $R^*$ in FRFL so as to maximize the training performance. As a first step, we will design a (learning-based) approach that dynamically returns the optimal choice for $R^*$ as a function of the dynamics of the environment in which the participating clients operate.
	
	
	\bibliographystyle{IEEEtran}
	\bibliography{biblio.bib}

\end{document}